\pdfoutput=1

\documentclass[11pt]{article}

\usepackage[final]{acl}
\usepackage{ragged2e} 
\usepackage{times}
\usepackage{latexsym}

\usepackage[T1]{fontenc}

\usepackage[utf8]{inputenc}

\usepackage{microtype}

\usepackage{inconsolata}

\usepackage{graphicx}

\usepackage{times}
\usepackage{helvet}
\usepackage{courier}
\usepackage{graphicx}
\usepackage{amsmath}
\usepackage{amsfonts}
\usepackage{amssymb}
\usepackage{color}
\usepackage{xcolor}
\usepackage{soul}
\usepackage{comment}
\usepackage{multirow}
\usepackage{booktabs}
\usepackage{todonotes}
\usepackage{subcaption}
\usepackage{tcolorbox}
\usepackage{enumitem}
\usepackage{colortbl}

\definecolor{darkgreen}{rgb}{0,0.5,0}

\title{Incongruence Identification in Eyewitness Testimony}

\author{
    Akshara Nair \quad Zeba Afroz \quad Md Shad Akhtar \\
    IIIT-Delhi, New Delhi, India \\
    \texttt{\{akshara22008, zebaa, shad.akhtar\}@iiitd.ac.in}
}

\newcommand{\dataset}{{\tt MIND}}
\newcommand{\model}{{\tt INTEND}}

\begin{document}
\maketitle
\begin{abstract}
Incongruence detection in eyewitness' narratives is critical for understanding the reliability of testimonies, yet traditional approaches often fail to address the nuanced inconsistencies inherent in such accounts. In this paper, we introduce a novel task of incongruence detection in eyewitness testimonies. Given a pair of testimonies consisting of multiple pairs of questions and answers by two subjects, we identify contextually-related incongruence between the two subjects. We also mark the span of incongruence in the utterances. To achieve this, we develop \dataset\ (\textbf{M}ult\textbf{I}-eyewit\textbf{N}ess \textbf{D}eception) -- a comprehensive dataset consisting of 2,979 pairs of contextually related answers designed to capture both explicit and implicit contradictions. Further, we propose an \textbf{IN}struction-\textbf{T}un\textbf{E}d i\textbf{N}congruity \textbf{D}etection framework based on the $6Ws$ and Multi-Hop reasoning approach, aka. \model. Drawing from investigative techniques, \model\ addresses the task as a cloze-style problem, concentrating on the \textit{who}, \textit{what}, \textit{when}, \textit{where}, and \textit{why} aspects of the context. Our findings show that prompt tuning, especially when utilizing our proposed framework, enhances the detection of incongruences by a margin of $+5.63\%$. 
We compare our approach with multiple fine-tuning and prompt-tuning techniques on MLMs and LLMs. Empirical results demonstrate convincing performance improvements in F1-score over fine-tuned and regular prompt-tuned techniques, highlighting the effectiveness of our approach.
\end{abstract}

\section{Introduction}

Eyewitness testimony has long been seen as an essential component of judicial and investigative processes, providing human perspectives that can reveal the truth about an occurrence. However, when many eyewitnesses submit statements, differences or contradictions, known as inter-eyewitness incongruence, may arise. Moreover, inter-eyewitness incongruence can occur for various point of references, such as the order of events, descriptions of the people involved, specifics of the activities committed, etc. These inconsistencies call into question the entire knowledge of the event and raise serious doubts about the testimonies' trustworthiness and credibility.
Historically, research on eyewitness testimony has concentrated chiefly on the linguistic examination of individual narratives \cite{doggett-cantarero-2016-identifying,sola2023analysing,Rachel}, often limiting the analysis to one perspective and interaction with other testimonies has been rarely explored. These approaches limit our knowledge of how inconsistencies occur and how they affect the overall narrative of an event and its key details.

\begin{table}[t]
    \centering
    \resizebox{\columnwidth}{!}{
    \begin{tabular}{p{20em}cp{20em}}
         \multicolumn{3}{l}{\textbf{Context:} \textit{What did the girl with black curly hair do after reaching the house of a man in the black jacket?}}  \\ \toprule 
         \textbf{Testimony $T1$:} So, when she reached [\textcolor{red!90}{\textit{Frederick's house, she knocked, then she hid.  When Frederick came out, she snapped his neck and killed him}}]$I_1$.  Then she entered the house to check for other members. Going to the first floor, she knocked on a door, a man came out, and [\textcolor{blue!90}{\textit{she threw him off the roof, causing him to die}}]$I_2$.  Then she entered the room from which the man came, where she found a woman lying on the bed, whom she proceeded to shoot. & & \textbf{Testimony $T2$:} When she get there, she found a boy so she thought he might be the son of that person so  [\textcolor{red!90}{\textit{she didn’t want to kill him but he attacked on that curly hair girl then in response to that curly hair girl like killed him}}]$I_1$ and then she entered in that person's house and she knock the door in which that person was there and and hide in a corner and then that man cried that who is this but she didn't reply and then [\textcolor{blue!90}{\textit{when he opened the door she killed and she attacked that man with the knife and killed him}}]$I_2$. \\ \toprule
         
         \textbf{Identified Incongruences:} \\
         \multicolumn{3}{l}{\multirow{3}{43.5em}{ \begin{itemize}
             \item \textcolor{red}{$I_1$:} $T1$ and $T2$ describe different sequences and outcomes for the girl’s actions upon arriving at the house., In T1, she kills someone inside the house after knocking on the door, while in T2, she kills Frederick immediately after he opens the door and before checking for other members.
              \item \textcolor{blue}{$I_2$:} $T1$ describes a knife attack leading to death, whereas T2 describes a man falling from the roof. \end{itemize}}} \\ \\
              \\ \\ \\ \\

    \end{tabular}}
    \caption{For a given context, $T_1$ and $T_2$ are testimonies from two witnesses taken in isolation. $I_1$ and $I_2$ are the spans of identified incongruencies.}
    \label{tab:example}
    \vspace{-1mm}
\end{table}

Traditional approaches for contradiction detection have relied on explicit linguistic cues like antonym, negation, and quantitative incompatibilities to discover contradictions. For instance, the detection of contradictions often involves recognizing antonyms (e.g., `full' vs. `empty'), the presence of negation (e.g., `is' vs. `is not'), or discrepancies in numerical information (e.g., `three' vs. `five') \cite{stanfordNLP,CCI}. However, these algorithms fail to detect subtle and implicit errors that are frequent in eyewitness testimonies. 
Moreover, most existing datasets and algorithms classify entire pair of sentence as either contraditing or not, but they do not identify the exact spans or segments of text where the contradiction occurs. The absence of adequate reasoning for these incongruences is a critical challenge that needs to be fully addressed. This shortcoming is primarily due to the lack of ground truth data that pinpoints the specific spans where contradictions exist. Without this level of detail, it is difficult to identify the nature of the contradiction and effectively address it, especially in the context of analyzing multiple witness statements.
\vspace{-0.26cm}
\paragraph{Motivation: }
Consistent testimonies from various eyewitnesses imply dependability and variations may indicate dishonesty or attempts to mislead investigators. Our attempt to explore the task of  detecting incongruence and providing appropriate reasoning in terms of textual for contradictions within testimonies is motivated by the critical importance in exposing potential dissimulation and mendacity. By pinpointing the exact segments where contradictions occur, span extraction provides irrefutable evidence and reason of these inconsistencies. This method transforms abstract incongruence into concrete, verifiable data, allowing for a more precise and informed evaluation of testimonies.
Table \ref{tab:example} exemplifies how two testimonies, $T1$ and $T2$, offer differing accounts of the same event. The discrepancies, $I1$ and $I2$, reveal variations in the sequence and outcomes of events, such as differing accounts of killing inside or outside the house and whether a man is killed by a knife or a fall. Pinpointing these nuanced discrepancies provide a clearer understanding of the conflicting narratives.


To this end, we develop \dataset, a multi-eyewitness deception detection dataset. It contains $389$ eyewitness testimonial statements related to $149$ events, each event observed by one, two, or three eyewitnesses, and $2,979$ pairs of testimonies with linked context-answer pairs, where each context is a predefined question about a subevent and the answers are the eyewitnesses' statements related to that context.
Further to enhance the reliability and accuracy of eyewitness testimonies, we propose an instruction-tuned incongruity detection framework, \model. Unlike conventional contradiction detection systems designed for short, factual texts, \model\ focuses on identifying discrepancies within complex narratives that address similar or related topics. 
Traditionally, the 5W approach \footnote{https://en.wikipedia.org/wiki/Five\_Ws.} -- focusing on who, what, when, where, and why -- has been a cornerstone of investigative techniques \cite{mott1942trends}. Building on this foundation, we introduce an expanded 6Ws framework, emphasizing on who (identity), what (action), what (object), when (timeline), where (location), and why (reason). By analyzing agreements, contradictions, and omissions within these six areas, we hypothesize that this method will reveal subtle yet significant differences that may not be apparent when using traditional approaches.



For the reasoning task, we identify the exact spans where they occur in a pair of testimonies and identify their incongruence alignment. We adopt a multi-hop strategy to extract subtle intricacies of incongruency in three steps, i.e., \textit{identify fine-grained key details}, \textit{infer incongruent reasons}, and \textit{extract conflicting spans}.
Our evaluation on \dataset\ demonstrate significant improvements in F1-scores against various MLMs and LLMs for both tasks.


\subsection{Contributions:} We summarize our contributions as:

  

\begin{itemize}[itemsep=0pt, ]
    \item We introduce a \textbf{novel task} of incongruence detection and reasoning task aimed at extracting contradicting statements between two testimonies.
    \item We develop \dataset\ -- a comprehensive \textbf{dataset} to train and evaluate the incongruence detection and reasoning tasks.
    \item We propose \model\footnote{Supplementary accompanies the code.}, a novel instruction-tuned framework supplemented with the $6Ws$ and multi-hop strategies for the detection and reasoning tasks, respectively.
    
\end{itemize}

\section{Related Work}
\paragraph{Existing Datasets:} One of the key areas where detecting incongruence is crucial is during interrogations, where individuals may attempt to deceive by presenting statements that contradict established evidence, facts, or the testimonies of other witnesses. Deception research primarly uses two types of dataset. Real-life and Mock. The Real Life Trial dataset \cite{perez2015verbal}\,includes misleading and honest footage extracted from courtroom footage while Mock datasets , generated under controlled conditions, include the work by \cite{CSC} with 32 hours of deceptive speech.
The Bag-of-Lies dataset by\cite{Bag-of-lies}, contains visual, audio, EEG, and eye gaze data collected in real-life situations utilizing standard mobile devices and sensors.
\vspace{-0.3cm}
\paragraph{Contradiction Detection:} A large portion of contradiction detection research falls under the scope of Natural Language Inference where the objective is to determine whether a hypothesis is entailing, contradicting or undetermined given a premise \cite{stanford-nlp-manning}. Early efforts focussed on sentence-level entailment, such as the work of \cite{Khot_Sabharwal_Clark_2018} \cite{schuster-etal-2022-stretching} extended NLI to longer documents with SeNtLI, through it retained sentence-level decomposition, which restricted context preservation.
However, no existing work effectively detects contradictions in personal narratives involving differing perspective.
\vspace{-0.3cm}
\paragraph{Large Language Models: } LLMs have recently gained popularity in various natural language processing tasks \cite{gpt}. Open-source models like Llama \cite{dubey2024llama} and Mistral \cite{jiang2023mistral} excel in handling complex tasks through natural language instructions. Instruction tuning, which fine-tunes LLMs using datasets paired with specific instructions, has also become widespread (Jiao et al., 2023a; Wang et al., 2023b; Zhang et al., 2023; Cheng et al., 2023), enabling the models to make task-specific predictions.

Additionally, recent works have started to leverage the common sense understanding of these models\cite{paranjape-etal-2021-prompting}. Research by \cite{fei2023reasoning,jiang-etal-2022-understanding} has demonstrated the value of multi-hop reasoning in enhancing model performance, especially in tasks that demand deep contextual understanding. The reasoning ability is essential for semantic understanding task like contradiction detection, where the model must grasp subtle nuances and approach the problem by mimicking human-like, step-by-step thinking. To the best of our knowledge no existing work detects contradictions in personal narratives involving multiple events or differing perspectives from various individuals, and all current approaches follow a binary method. In contrast, \model\ distinguishes itself by operating at both the binary and span levels, addressing the gap in analyzing complex, multi-perspective datasets.

\section{Dataset}
As discussed earlier, existing datasets on deception detection \cite{perez2015verbal,CSC,Bag-of-lies} do not support the proposed task of incongruence detection for eyewitness testimonies. Moreover, these datasets frequently disregard several perspectives on a particular matter, resulting in potential prejudices and an inadequate comprehension of deceitful conduct. To this end, we develop \dataset\ in two stages. At first, we adopted wizard-of-oz setup to collect testimonies in a controlled interrogation environment specifically designed to replicate real-life situations. Subsequently, we employ annotators to tag necessary labels. 

\subsection{Data Collection}
In this section, we lay out detailed process of the data collection phase.
\begin{itemize}
    \item \textbf{Event Selection:} Initially, we meticulously selected high-quality videos focusing majorly on crime-related topics such as, robbery, theft, murder, etc. We follow these guidelines to ensure a comprehensive representation of criminal contexts:
        \begin{itemize}
            \item \textbf{Realism:} We compile films that primarily depict  criminal incidents or situations. A few non-crime scenarios have been included as well, but they are kept to a minimum to maintain focus.
            \item \textbf{Diversity:} The dataset encompasses a diverse array of criminal scenarios, specifically designed to challenge witnesses and elicit scenario-specific deception, enabling the analysis of how individuals fabricate in various contexts.
        \end{itemize} 
    
    \item \textbf{Preparing witness:} Next, we show these videos to human actors (aka. witness) about a particular situation. This step mimics the real-life experience of a person who witnessed the event. We allowed a cooling-off period (15 mins to a few days) to witness before the interrogation. 
    
    \item \textbf{Controlled Interrogation:} During interrogation, witnesses are asked to recollect their observations of the event and record their testimonies as response to the interrogator's question. Interrogator's questions encompass both direct inquiries on the witnessed events and subsequent assessments to gauge the coherence and precision of the answers. Meanwhile, some of the witnesses are instructed to occasionally deceit at their own discretion -- the interrogator is aware of what event has happened (e.g., murder) but they are unaware of the details of the events, such as, suspect, timeline, equipment used for the crime, etc. The various manifestations of deceit encompass. 
        \begin{itemize}
            \item \textbf{Concealment:} Intentional omission or downplaying of details about the perpetrator, activities, or setting by a witness.
            \item \textbf{Fabrication:} Deliberate creation of false information.
            \item \textbf{Distortion:} Intentional alteration of details or characteristics of an observed event by a witness.
        \end{itemize}
    \item \textbf{Transcription:} Following the interview, we transcribe the testimonies into specified textual format.
    \item \textbf{Statistics:} Table \ref{tab:stat} reports statistics of \dataset. We collect $149$ videos across $17$ different events. The average duration of collected videos is approximately $12$ minutes with a minimum and maximum duration of $\sim4$ and $\sim20$ minutes, respectively. In total, we recorded $380$ testimonies with an average duration of $9.5$ minutes per testimony. In each testimony, there are $\sim15$ rounds of interrogator's questions and witness's response on average.  
\end{itemize}

    \begin{table}[t]
        \centering
        \resizebox{\columnwidth}{!}{
        \begin{tabular}{c|l|p{20em}}
           & \textbf{Metric} & \textbf{Value} \\ \toprule
           \multirow{6}{*}{\rotatebox{90}{Events}} & Number of Events & 149 \\
           & Avg. length of event videos & 12 minutes \\
           & Event topics & 17 [\textit{fraud}, \textit{kidnap}, \textit{robbery}, \textit{non-crime}, \textit{self-harm}, \textit{phishing}, \textit{murder}, \textit{gangsterism}, \textit{drug trafficking}, \textit{homicide}, \textit{accident}, \textit{racketeering}, \textit{fight}, \textit{bribery}, \textit{forgery}, \textit{bullying}, and \textit{sexual assault}]   \\ \midrule
           
           \multirow{3}{*}{\rotatebox{90}{Testimony}} & Number of Testimonies & 389 \\ 
           & Total Q\&A Pair Count & 6,000 \\ 
           & Interrogation Duration & 9 minutes 30 seconds (average) \\ \midrule

           \multirow{5}{*}{\rotatebox{90}{Incongruity}} & Number of unique Contexts & 1317 \\
           & Number of testimony pairs & 2979 \\
           & Number of Incongruent pairs  &  1850 \\ 
           & Number of Non-incongruent pairs & 1129 \\
           & Avg. number of Incongruent tokens & 10.34 \\ 
           
           \bottomrule
        \end{tabular}}
        \caption{Statistics of \dataset.}
        \label{tab:stat}
    \end{table}

\vspace{-0.3cm}
\subsection{Incongruence Annotations}
Following the data collection phase, we move onto preparing the dataset for incongruence identification in witness testimonies. For each event, we carefully drafted contextual questions resonating subtopics of the event. Subsequently, for each context, we identified portion of the testimonies  relevant to the context from a pair of testimonies, $T1$ and $T2$. Finally, annotators\footnote{We employ 4 annotators, 2 male and 2 female, who volunteered for the task. They belong to the age group of 25-30 years and are linguistics experts.} carefully annotate the inconsistent sections with contradicting information. This procedure entailed pinpointing textual spans reflecting the location of inconsistencies, providing a higher degree of information. To maintain uniformity and precision, we train annotators with an annotation guidelines and mandated them to adhere to them while annotating. We conduct multiple rounds of training sessions to ensure that annotators are adequately comfortable with the guidelines. Refer to Appendix~\ref{sec:appendixA} for detailed annotation guidelines.

\paragraph{Statistics:} Our dataset comprises 1,317 distinct contexts, encompassing a total of 2,979 testimony pairs. Within these pairs, 1,850 are identified as incongruent, while the remaining 1,129 are classified as non-incongruent. The average number of tokens in incongruent pairs is $10.34$, reflecting the typical span of identified incongruences within the testimonies. We summarize dataset statistics in Table \ref{tab:stat}.

\begin{figure}[t]
    \begin{tcolorbox}[width=\columnwidth,title=,colupper=black]\scriptsize
        \textbf{Instruction} = Compare and assess the statements of Witness A and Witness B regarding [specific aspect, e.g., identification, action, object, timeline, location, motive]. Identify agreements, contradictions, or missing information between their accounts. \\
        \textbf{Fill in the \texttt{mask} accordingly:}
        \begin{itemize}[itemsep=-2pt]
            \item[--] Witness A's \textbf{identification} of the person \texttt{[mask]} Witness B's identification.
            \item[--] Witness A's \textbf{described action} \texttt{[mask]} Witness B's description.
            \item[--] Witness A's \textbf{described object} \texttt{[mask]} Witness B's described object.
            \item[--] Witness A's reported \textbf{timeline} \texttt{[mask]} Witness B's reported timeline.
            \item[--] Witness A's reported \textbf{location} \texttt{[mask]} Witness B's reported location.
            \item[--] Witness A's reported \textbf{reason} \texttt{[mask]} Witness B's reported reason.
        \end{itemize}
        \textbf{\texttt{[mask]} options are:}
        \begin{itemize}[itemsep=-2pt]
            \item[ ] \textbf{agrees with:} Use when both testimonies provide consistent or additional details that align with each other.
            \item[ ] \textbf{contradict:} Use when testimonies directly conflict with each other.
            \item[ ] \textbf{is absent from:} Use when one testimony does not provide information on a detail covered by the other.
        \end{itemize}
    \end{tcolorbox}  
    \vspace{-3mm}
    \caption{An example of prompt with $6Ws$ instruction for incongruence detection.}
    \label{fig:prompt:6w}
\end{figure}

\section{Methodology}
In this section, we formulate our problem definitions and describe our proposed approach, \model. 

\subsection{Problem Formulation}
Given a triplet of context $C$ and a pair of testimonies, $T1$ and $T2$, we formulate following two tasks as follows:

\begin{itemize}
    \item \textbf{Incongruence Detection:} In this task, we identify the implied incongruence between $T1$ and $T2$ considering the context $C$ as a binary task, i.e., \textit{True} or \textit{False}. 

    \item \textbf{Incongruence Reasoning \& Alignment:} For each incongruent testimony pair, we generate reasoning for the incongruence by identifying contradiction in both testimonies and extract spans of text $[p_1, p_2, \cdots, p_n] \in T_1$ and $[q_1, q_2, \cdots, q_k] \in T_2$ that reveal contradictory information.     
\end{itemize}

\subsection{Proposed Methodology:  \model}
In this section, we lay out the details of \model. Our method combines instruction tuning along with the novel $6Ws$ approach and a multi-hop reasoning framework for the identification and reasoning tasks, respectively.

\paragraph{Incongruence Detection:} The $6W$ prompt, inspired by \citet{mott1942trends}, is a structured strategy for comparing two testimonies by evaluating certain features known as the ``\textit{6 W-type questions}". In particular, it strives to reveal the following details: 
\textbf{Identity: }\textit{who was the subject?}; \textbf{Action:} \textit{what action did they do?}; \textbf{Object:} \textit{what object was used?}; \textbf{Location:} \textit{where it was done?}; \textbf{Time:} \textit{when did it happen?}; and \textbf{Reason:} \textit{why they did it?}. 

This method seeks to find \textit{consistency}, \textit{discrepancies}, or \textit{absences} in the details offered by witnesses. In our proposed method, we leverages the nuances of these $6W$ parameters in prompt. By incorporating these factors, we hypothesize that the prompt construction would direct the model's attention to specific areas of the testimony, allowing for a more detailed comparison. For each $W$ in $6Ws$, we prompt the model if ``witness A's details \texttt{[mask]} witness B's details, where \texttt{[mask]} is defined as one of the three labels: `\textit{agrees}', `\textit{contradict}', or `\textit{absent}'. The label \textit{agrees} is expected when both testimonies provide consistent or additional details that complement each other, whereas, the label \textit{contradict} is used when testimonies directly contradict each other. In cases where one testimony lacks information on a topic covered by the other is labelled as \textit{absent}. These labels instruct the model to compare each component of the testimonies independently, allowing for more precise detection of contradictions or agreements between the witnesses' claims. An instance of a prompt with $6Ws$ statements is shown in Figure \ref{fig:prompt:6w}. 

\begin{table*}[ht]
    \centering
    \resizebox{\textwidth}{!}{
    \begin{tabular}{|lp{62em}|}
    \toprule
        \rowcolor{red!30} \multicolumn{2}{c}{\bf Hop 1 -- Identifying Fine-grained Key Details} \\ \midrule
        \bf Prompt: & Given the question and answers provided by two testimonies, analyze and summarize the key points described in each response relative to the question.  \\
        & \textbf{Question:} Describe the man to whom the bag was handed over to.  \\
        & \textbf{Witness A:} Yeah, he's wearing a shirt, and he's a fat guy. He's a long-haired man standing in the field, and it's a dark night. He was waiting there patiently smoking a cigarette near a trash can.  \\
        & \textbf{Witness B:} So the man was middle aged and he has some beard, and he was wearing a green jacket. And he was standing there smoking. And yeah, that's what I remember. So it was a random spot near a trash can. And as he was waiting there, he was quite frustrated. And he was waiting for quite some long time as he said that it's been hours waiting. And then he was so frustrated that he started throwing rocks, and he was kicking his car's tire. And then for whom he was waiting here, arrived. Date was never mentioned. But the time was it was a late night? Probably midnight? Yeah, that's what I remember. \\ 
        \bf Response: & \textbf{Physical Description:} Account A describes the man as fat, long-haired, and wearing a shirt. Account B describes him as middle-aged with a beard, wearing a green jacket.  \\
        & \textbf{Location:} Both accounts place the man near a trash can, but Account A specifies it's in a field.  \\
        & \textbf{Behavior:} Account A mentions the man was patiently smoking. Account B describes him as initially frustrated, throwing rocks and kicking his car's tire.  \\ \bottomrule
        \toprule
        \rowcolor{green!30} \multicolumn{2}{|c|}{\bf Hop 2 -- Inferring Incongruence Reason} \\  \midrule
        \bf Prompt: & Using the summary of the key points described in each response relative to the question, use common sense and determine if there is a direct conflict between any details in Accounts A and B. If there is no conflict, return ``\textit{No conflict detected}". If there is a conflict, explain why the details are inconsistent. \\
        & \textbf{Physical Description:} Account A describes the man as fat, long-haired, and wearing a shirt. Account B describes $\cdots$. \\
        & \textbf{Location:} Both accounts place the man near a trash can, but Account A specifies it's in a field.  \\
        & \textbf{Behavior:} Account A mentions the man was patiently smoking. Account B describes him as initially frustrated $\cdots$. \\ 
        \bf Response: & \textbf{Clothing Description:} Account A describes the man as wearing a shirt. Account B mentions the man wearing a green jacket. \textbf{\em Potential Conflict:} This detail is conflicting because a person cannot simultaneously wear just a shirt and a green jacket. It suggests inconsistency regarding what the man was wearing at the time of the incident.  \\
        & \textbf{Behavior:} Account A portrays the man as patiently smoking. Account B describes the man as initially frustrated, throwing rocks and kicking his car's tire. \textbf{\em Potential Conflict:} This is a direct conflict because Account A portrays the man as calm and patient, engaging in a mundane activity (smoking), while Account B portrays him as visibly frustrated and engaging in more intense actions (throwing rocks, kicking car tire). These behaviors are contradictory in terms of the emotional state and actions attributed to the man during the same situation.  \\
        & \textbf{Time Frame:} Account B specifies it was late at night, possibly around midnight. Account A does not mention a specific time frame, only stating it was a dark night. \textbf{\em Potential Conflict:} While not a direct contradiction, the lack of a specific time frame in Account A could conflict with the specificity of Account B regarding the time of the incident. \\ \bottomrule 
        \toprule
        \rowcolor{blue!30} \multicolumn{2}{|c|}{\bf Hop 3 -- Extracting Conflicting Spans} \\ \midrule
        \bf Prompt: & Given an explanation, a question and answers provided by two testimonies you are tasked with extracting and analyzing incongruent segments of text between Witness A and Witness B using the explanation. \\
        & \textbf{Question:} Describe the man to whom the bag was handed over to.  \\
        & \textbf{Witness A:} Yeah, he's wearing a shirt, and he's a fat guy $\cdots$. \\
        & \textbf{Witness B:} So the man was middle aged and he has some beard $\cdots$. \\
        & \textbf{Explanation:} \%(from hop 2)\% \\
        & $\quad$ \textbf{Clothing Description:} Account A describes$\cdots$. \textbf{\em Potential Conflict:} This detail is conflicting because $\cdots$.  \\
        & $\quad$ \textbf{Behavior:} Account A portrays the man as $\cdots$. \textbf{\em Potential Conflict:} This is a direct conflict because $\cdots$  \\
        & $\quad$ \textbf{Time Frame:} Account B specifies it was $\cdots$. \textbf{\em Potential Conflict:} While not a direct contradiction, the lack of a $\cdots$. \\
        \bf Response: & \textbf{Contradiction 1:} \textcolor{blue!90}{["\textit{he's wearing a shirt}" from Witness A]} \textcolor{red!90}{\textbf{\em contradicts}} \textcolor{darkgreen}{["\textit{he was wearing a green jacket}" from Witness B] }\\ 
        & \textbf{Contradiction 2:} \textcolor{blue!90}{["\textit{He was waiting there patiently smoking a cigarette near a trash can}" from Witness A]} \textcolor{red!90}{\textbf{\em contradicts}} \textcolor{darkgreen}{["\textit{he was so frustrated that he started throwing rocks, and he was kicking his car's tire}" from Witness B]} \\ \bottomrule
    \end{tabular}}
    \caption{A three-hop approach for incongrunce reasoning and alignment.}
    \label{tab:multi-hop}
\end{table*}

\paragraph{Incongruent Reasoning and Alignment:} Testimonies frequently contain intricate details and subtle discrepancies that may not be well represented in a small training set commonly utilized for few-shot learning. On the other hand, fine-tuning necessitates a large amount of labeled data, which can be expensive and time-consuming to obtain, especially given the variety and nuanced inconsistencies found in testimony.

To overcome these issues, we created a \textbf{multi-hop reasoning} framework based on the work of \citet{fei2023reasoning,chain-of-thought}. This framework utilizes the ability of common sense understanding and improves the extraction of contradicting spans by combining information from numerous segments of text and repeatedly linking and reasoning across them. This enables the model to gain a thorough understanding of the context around contradictions, resulting in more accurate detection and extraction of incongruent spans. Additionally, we experimented with varying the number of reasoning hops to assess their impact on performance. Detailed results provided in Appendix~\ref{sec:appendixE}.



We adopt three-hop strategy with different prompts to gradually improve \model's reasoning capabilities. This organized approach addresses the intricacies of the problem by gradually refining the model's understanding.

\begin{itemize}
    \item \underline{Identifying Fine-grained Key Details}: The first prompt focuses on extracting fine-grained critical facts from the two accounts that relate to the supplied inquiry. This stage ensures that the model correctly understands the precise information presented in each testimony, establishing the groundwork for later reasoning.
    \item \underline{Incongruence Reasoning}: The second task requires using common sense and logical reasoning to uncover probable discrepancies between the accounts. By examining the context and content of the testimony, the model can uncover small inconsistencies that may not be immediately obvious.
    \item \underline{Extracting Conflicting Spans}: The third task focuses on extracting specific text spans that have contradictions using the rationale from the previous stage. This stage entails identifying the specific segments that contradict each other, guided by the previous conclusions, and guaranteeing precision in contradiction identification.
\end{itemize}

This incremental reasoning approach mimics human-like behaviour, which progresses from simpler to more complicated problems, simplifies span extraction while effectively tackling the challenges of direct incongruence prediction. By exploiting \model's multi-hop capacity to integrate and reason through various layers of information, we obtain inconsistent spans more precisely and reliably. Table \ref{tab:multi-hop} depicts a three-hop strategy to \textit{identify the key-details}, \textit{infer the reasoning}, and \textit{extract the incongruent span}.

\section{Experiments, Results, and Analysis}

This section presents the experimental setup, the results obtained, and a detailed analysis of our findings.

\begin{table*}[t]
    \centering

    \begin{subtable}{0.56\textwidth}
        \resizebox{\textwidth}{!}{
        \begin{tabular}{l|ccc|ccc|ccc}
            \hline
            & \multicolumn{3}{c|}{Fine-tuned MLM} & \multicolumn{3}{c|}{MLM -- Regular prompts} & \multicolumn{3}{c}{MLM -- Question prompts}  \\
            \hline
            \textbf{Model} & \textbf{Pre} & \textbf{Rec} & \textbf{F1} & \textbf{Pre} & \textbf{Rec} & \textbf{F1} & \textbf{Pre} & \textbf{Rec} & \textbf{F1} \\
            \hline
            \bf Longformer & 0.59 & 0.59 & 0.59 & 0.63 & 0.63 & 0.62 & 0.66 & 0.63 & 0.63 \\
            \bf Big-Bird & 0.63 & 0.62 & 0.62  & 0.64 & 0.64 & 0.64 & 0.65 & 0.65 & 0.64\\
            \bf Bird-MNLI & 0.64 & 0.63 & 0.63 & 0.64 & 0.62 & 0.63 & 0.65 & 0.63 & 0.63 \\
            \hline
        \end{tabular}}
        \caption{Masked Language Models.}
        \label{table:performance:PLM}
    \end{subtable}
\hfill
    \begin{subtable}{0.42\textwidth}
    \resizebox{\textwidth}{!}{
        \begin{tabular}{l|ccc|ccc}
            \hline
        & \multicolumn{3}{c|}{LLM -- Question prompts} & \multicolumn{3}{c}{\model\ -- $6Ws$ prompts} \\
            \cline{2-7}
            \textbf{Model} & \textbf{Pre} & \textbf{Rec} & \textbf{F1} & \textbf{Pre} & \textbf{Rec} & \textbf{F1} \\
            \hline
            \bf Llama-3 [8B] & 0.66 & 0.68& 0.67 & 0.62  & \bf 0.89 & 0.73 \\
            \bf Gemma [9B] & 0.62 & 0.69 & 0.65 & 0.64 &  0.76& 0.70 \\
            \bf Mistral [7B] & 0.63 & 0.83 & 0.71 & \bf 0.65  & 0.87 & \textbf{0.75} \\
            \bf Qwen [7B] & 0.61 & 0.75 & 0.67 & 0.64 & 0.74 & 0.69 \\
            \hline
        \end{tabular}}
        \caption{Instruction-tuned Large Language Models.}
        \label{table:performance:LLM}
    \end{subtable}
    \vspace{-3mm}
    \caption{Performance evaluation of incongruence detection on \dataset\ using different tuning approaches.}
    \label{table:performance}
\end{table*}

\begin{table*}[!ht]
\centering
\resizebox{0.88\textwidth}{!}
{
\begin{tabular}{l|l|ccc|ccc|ccc}
\hline
& & \multicolumn{6}{c|}{\textbf{Span Identification}} & \multicolumn{3}{c}{\textbf{Incongruence Alignment}} \\
\cline{3-10}

\textbf{Setup} & \textbf{Model} & \multicolumn{3}{c|}{\textbf{Testimony $T1$}} & \multicolumn{3}{c|}{\textbf{Testimony $T2$}} & \multicolumn{3}{c}{\textbf{between $T1$ \& $T2$}} \\
\cline{3-11}
 & & \textbf{Pre} & \textbf{Rec} & \textbf{F1}  & \textbf{Pre} & \textbf{Rec} & \textbf{F1} &  \textbf{Pre} & \textbf{Rec} & \textbf{F1} \\
\hline
\multirow{4}{*}{\rotatebox{0}{\bf Few-shot}} & \textbf{LLAMA-3 [8B]} & 0.434 & 0.536 & 0.431 &  0.456 & 0.519 & 0.436 &  0.456 & 0.577 & 0.450 \\
& \textbf{QWEN-2 [7B]} & \bf 0.513 & 0.482 & 0.447  & \bf 0.527 & 0.455 & 0.441  & 0.536 & 0.508 & 0.466 \\
& \textbf{MISTRAL [7B]} & 0.421 & 0.461 & 0.404  & 0.428 & 0.453 & 0.404  & 0.467 & 0.509 & 0.436 \\
& \textbf{GEMMA [9B]} & 0.405 & 0.569 & 0.425  & 0.433 & 0.550 & 0.439  & 0.425 & 0.614 & 0.451 \\
\hline
\hline
\multirow{4}{*}{\rotatebox{0}{\bf \model\ -- Multi-hop}} & \textbf{LLAMA-3 [8B]} & 0.496 & \bf 0.583 & \bf 0.490  & 0.515 & \bf 0.579 & \bf 0.498  & 0.519 & \bf 0.621 & \bf 0.512 \\
& \textbf{QWEN-2 [7B]} & 0.505 & 0.381 & 0.395  & 0.518 & 0.369 & 0.390  & \bf 0.557 & 0.423 & 0.431 \\
& \textbf{MISTRAL [7B]} & 0.466 & 0.503 & 0.453  & 0.479 & 0.488 & 0.454  & 0.525 & 0.530 & 0.482 \\
& \textbf{GEMMA [9B]} & \bf 0.513 & 0.564 & \bf 0.490  & 0.509 & 0.573 & 0.489  & 0.523 & 0.611 & 0.507 \\
\hline

\end{tabular}}
\vspace{-3mm}
\caption{Performance evaluation of Incongruence span identification on \dataset\ using different tuning approaches.}
\label{table:performance:span}
\end{table*}

\subsection{Baselines}
We adopt three classes of baseline methods for comparison: 

\subsubsection{\textbf{a)} Pre-trained MLM with finetuning:}
We fine-tune three transformer models -- Longformer \cite{beltagy2020longformer}, Big-Bird \cite{zaheer2020big}, and Big-Bird-MNLI fine-tuned on the MNLI dataset. We select these baseline for their capacity to handle larger context of up to $4K$ tokens as the max combined input size is $\sim600$ tokens -- surpassing the $512$ token limit of standard models like BERT, RoBERTa, etc.

\subsubsection{\textbf{b)} Pre-trained MLM with prompt tuning:}
Inspired by the works of \cite{prompt_liu,schick2021exploitingclozequestionsshot}, we develop and fine-tune MLM models using two distinct prompt templates using \texttt{[input]} defined as a question-answer triplet, \textbf{Question Prompt:} 
     ``\texttt{[input]} Is there a direct contradiction between the statements made by Witness A and Witness B? \texttt{[mask]}." and \textbf{Regular Prompt:} 
     ``\texttt{[input]} A? \texttt{[mask]} B."

    

\subsubsection{\textbf{c)} Pre-trained LLM with instruction tuning:}
We utilized four large language models (LLMs) for our experiments: LLama-3 \cite{dubey2024llama}, Mistral \cite{jiang2023mistral}, Gemma \cite{team2024gemma}, and Qwen \cite{bai2023qwen}. We instruction-tuned each one using the question prompt described above as the instructions.

\subsection{Experimental Settings}
We perform experiments on \dataset\ and use 65:35 split
to create train (1938), and test (1049) sets. 
For 6W instruction tuning, we annotate 151 samples with 6W template where the [mask] in the template(refer to Figure \ref{fig:prompt:6w}) is replaced with correct label(agrees, contradicts, absent) and use them as training set. We employ classical evaluation metrics, such as, precision, recall, and F1-score for both tasks. Additionally, we report \textit{coverage} for the span extraction task, which computes the overlaps between the predicted span considering the given testimony. This is crucial as \model\ employs LLMs to generate the spans instead of extracting it. Thus, we calculate the ratio of tokens in the predicted span belonging to the testimony or otherwise. Hence, we define \textit{coverage} as: 
$$ \text{Coverage} = \frac{|\text{Predicted Span} \cap \text{Testimony}|}{|\text{Predicted Span}|}
$$
Implementation details are furnished in the Appendix~\ref{sec:appendixB}. 

\begin{table*}[ht]
\centering

\resizebox{\textwidth}{!}{
\begin{tabular}{lcp{25em}p{25em}p{20em}}
\toprule
\multirow{1}{*}{\bf Model} & \multirow{1}{*}{\bf Hop} & \multicolumn{1}{c}{\bf Incongruent Span in $T1$ } & \multicolumn{1}{c}{\bf Incongruent Span in $T2$ } & \multirow{1}{*}{\bf Remarks} \\ 
\midrule \midrule
\rowcolor{cyan!60} \multicolumn{4}{l}{\textbf{Context:} How did Conner Help Shaun?} & \\ \hline
Gold & & Conner gave a \textcolor{blue!90}{\emph{\textbf{gave him the money and like other facilities.}}} and and told him to go for underground for like three months that was the help. & so conner \textcolor{blue!90}{\emph{\textbf{provided shaun with the contact of a man named Broadridge.}}} He said that Broadridge will be able to help you deal with the italian gang. &  Actual Incongruent pair  \\ \midrule
                   
\multirow{5}{*}{\model} & 1 & \multicolumn{2}{l}{\multirow{1}{60em}{Overall, while the two accounts differ in the specifics of how Conner helped Shaun, they both suggest that Conner played a role in supporting Shaun in some way, whether it was providing financial assistance, logistical support, or connecting him with someone who could help him with his problems.}} & \multirow{3}{20em}{The models failed to accurately capture the specific differences in how Conner assisted Shaun in 2\textsuperscript{nd} hop, mistakenly treating both events as the same rather than recognizing them as contradictory.} \\ \\

& 2 & \multicolumn{2}{l}{\multirow{1}{60em}{The key difference between the two accounts is the nature of the assistance provided by Conner. However, both accounts suggest that Conner helped Shaun in some way, and there is no direct conflict between the details provided.}} &  \multirow{3}{20em}{} \\ \\

& 3 & \textcolor{red!90}{\emph{\textbf{No contradiction}}} & 
 & \multirow{3}{20em}{} \\ \midrule{}

Baseline & & \textcolor{red!90}{\emph{\textbf{Conner}}} \textcolor{darkgreen!90}{\emph{\textbf{gave a gave him the money and like other facilities}}} \textcolor{red!90}{\emph{\textbf{and and told him to go for underground for like three months that was the help.}}} & \textcolor{red!90}{\emph{\textbf{Conner}}} \textcolor{darkgreen!90}{\emph{\textbf{provided shaun with the contact of a man named Broadridge.}}} & \multirow{3}{20em}{Model predicted contradiction, But with incorrect boundaries} \\

\bottomrule
\end{tabular}}
\vspace{-3mm}
\caption{Error Analysis. Gold spans are highlighted in blue. Red text denotes incorrect spans by our model \model\ and baseline (Llama-3 Few-Shot), while the green text highlights the correct spans for incongruent testimony pairs.
}
\label{tab:error}
\end{table*}





\begin{table}[t]
\centering

\resizebox{\columnwidth}{!}{
\begin{tabular}{l|ccc|ccc|ccc|ccc}
\multirow{2}{*}{Model} & \multicolumn{3}{c}{Clarity} & \multicolumn{3}{c}{Logical Exclusivity} & \multicolumn{3}{c}{Context Relevance} & \multicolumn{3}{c}{Coverage} \\ \cline{2-13}
& Good & Fair & Poor & Good & Fair & Poor & Good & Fair & Poor & Good & Fair & Poor\\
\midrule
FS & $33.4$  & $28.7$ & $37.8$  & $26.4$ & $35.0$ & $38.5$ & $36.6$ & $39.3$ & $24.0$ & $30.4$ & $39.7$ & $29.7$ \\
\midrule
MH & $52.8$  & $27.8$ & $19.2$ & $45.9$ & $35.8$ & $18.1$ & $57.2$ & $35.5$ & $7.2$ & $57.5$ & $34.4$ & $7.9$ \\ \bottomrule
\end{tabular}}
\caption{Human evaluation on Llama-3 in few-shot (FS) and \model's multi-hop (MH) setups. Numbers indicate the percentage of times evaluators preferred the respective model’s output across the defined categories}
\vspace{-1em}
\label{tab:Human-eval}
\end{table}
\vspace{-0.3cm}
\subsection{Results}
We report precision, recall, and F1-scores for incongruence detection in Table \ref{table:performance}. In Table \ref{table:performance:PLM}, we list comparative results in a fine-tuned and two prompt-tuned settings for three MLM models, i.e., Longformer, Big-Bird, and Bird-MNLI. Though we observe minimal variations across different tuning strategies, Big-Bird reports the best F1-score of $0.64$ in the prompt-tuned settings. In comparison, Longformer yields the most significant improvement when moving from fine-tuning (F1-score of $0.59$) to question-prompt-tuning (F1-score of $0.63$), highlighting its potential for improved performance with more tailored prompts. Moreover, question-prompt-tuned models score relatively better than the other two settings for all models.  

Similarly, we report the results for instruction-tuned LLMs including our proposed \model\ model in Table \ref{table:performance:LLM}. With question-prompts, Mistral [7B] reports the best F1-score of $0.71$. In comparison, other LLMs, such as, Llama-3 [8B], Qwen [7B], and Gemma [9B], report inferior F1-scores of $0.67$, $0.67$, and $0.65$, respectively. Overall, performances of LLMs with question prompts are better than the question-prompt-tuned MLM models. Finally, we report the performance of \model\ with $6Ws$ prompts. It is evident from Table \ref{table:performance} that \model\ demonstrates the best performance of $0.75$ F1-score with Mistal [7B] as the foundational model -- a clear improvement of $+5.63\%$ in F1-score against Mistal [7B] with question-prompts. Furthermore, we observe that \model\ yields improved performance for each LLM in the range of $[3-8\%]$ against the question-prompt strategy. This indicates the significance of constructing our instructions with $6Ws$ prompts.

Table \ref{table:performance:span} shows comparative results on the incongruence reasoing and alignment task. As mentioned earlier, we project the incongruence reasoning task as the span identification task through a generative framework. In the few-shot setup, QWEN-2 [7B] exhibits the best performance with an F1-score of $0.447$ in T1 and $0.441$ in T2 for the span identification task. It also yields the best F1-score of $0.466$ for the alignment task.
In contrast, the multi-hop approach significantly outperforms the few-shot baselines by a margin of $>9\%$ in F1-score for both span identification and alignment task.
Llama-3 demonstrates an F1-score of $0.49 (+9.63\%)$ on T1 and $0.49  (+12.92\%$) on T2 for the span identification task. In the alignment task as well, it records an F1-score of $0.51$, which $+9.87\%)$ improvement against the best baseline. Gemma, as the second-best model, also reports notable improvements. It achieves an F1-score of $0.49 (+9.63\%)$ on T1, $0.489 (+10.8\%)$ on T2, F1-score of $0.507 (+8.79\%)$ in the alignement task.
We also observe that the coverage improves for most of models in the multi-hop setup. It signifies that the generated spans increasingly aligned with the relevant testimonies. Additionally, we experiment with GPT-4o mini. See Appendix~\ref{sec:appendixF}.

\subsection{Error Analysis}

We investigate model's outcome for specific errors. Table \ref{tab:error} displays two randomly selected sample (due to brevity, we present other error cases in Appendix~\ref{sec:appendixD}) from the \dataset\ dataset along with its predicted and reference incongruent spans. We also incorporate the best-performing baseline, Llama-3(Few-Shot), for a fair comparison.

\subsection{Human Evaluation}

To conduct human evaluation, we randomly sample the outputs of 15 test set examples for both the baseline and multi-hop Llama-3 models. Thirty human evaluators rate the quality of the generated spans across four categories: \textit{clarity}, \textit{mutual exclusivity}, \textit{context relevance}, and \textit{coverage}, considering identified incongruencies. Each category has three rating levels: poor, fair, and good. Evaluators also assess whether all relevant contradictions are captured and whether unnecessary contradictions are included, choosing between two options: yes or no. This ensures the models' completeness and precision. For detailed evaluation criteria, refer to Appendix~\ref{sec:appendixC}. Table \ref{tab:Human-eval} shows the percentage of times evaluators preferred each model's outputs across the defined categories.

\section{Conclusion}

In this paper, we introduce a novel task of multi-eyewitness incongruence detection, focusing on identifying and extracting spans responsible for inconsistencies in two witness statements. To support this, we develop \dataset, a dataset pairing contexts with testimony responses and annotated incongruent spans. We also propose \model, an instruction-tuned framework based on 6Ws for detecting incongruence and a multi-hop reasoning approach for span extraction. Our evaluation shows \model\ outperforms baselines across various metrics, with human evaluations validating the quality of generated spans. We hope our contributions inspire further research in eyewitness testimony analysis and explainability.

\section{Limitations}

This study has two key limitations. First, the annotations used in our work relied on expert human annotators, ensuring precision in identifying nuanced incongruences. However, this reliance on manual processes highlights the need for semi-automated methods to enhance scalability without compromising accuracy. Second, the framework was evaluated primarily in the domain of crime-related narratives. While this ensured depth and relevance within the chosen context, extending the framework to other domains, such as medical or historical testimonies, would broaden its generalizability.


\section{Ethical Considerations}

This study adhered to strict ethical guidelines during data collection and analysis to ensure fairness, transparency, and respect for privacy. The testimonies used in the MIND dataset were collected under controlled conditions with informed consent from all participants. However, applying the proposed framework to real-world scenarios raises ethical concerns, such as the potential misuse of automated incongruence detection systems in legal or investigative contexts without proper oversight. It is crucial to emphasize that this framework is intended as an aid to human decision-making rather than a replacement. Future implementations must prioritize transparency, accountability, and ethical usage to mitigate biases and prevent unjust outcomes.


\bibliography{custom}

\appendix

\section{Annotation Guidelines}
\label{sec:appendixA}

\begin{table*}[!ht]
\centering
\resizebox{0.70\textwidth}{!}
{
\begin{tabular}{l|ccc|ccc|ccc}
\hline
& \multicolumn{6}{c|}{\textbf{Span Identification}} & \multicolumn{3}{c}{\textbf{Incongruence Alignment}} \\
\cline{2-7}

\textbf{Hops} & \multicolumn{3}{c|}{\textbf{Testimony $T1$}} & \multicolumn{3}{c|}{\textbf{Testimony $T2$}} & \multicolumn{3}{c}{\textbf{between $T1$ \& $T2$}} \\
\cline{2-10}
 & \textbf{Pre} & \textbf{Rec} & \textbf{F1}  & \textbf{Pre} & \textbf{Rec} & \textbf{F1}  & \textbf{Pre} & \textbf{Rec} & \textbf{F1} \\
\hline
\multirow{1}{*}{\rotatebox{0}{\bf One}}  & 0.459 & 0.467 & 0.421  & 0.449 & 0.447 & 0.406  & 0.488 & 0.509 & 0.446 \\

\hline

\multirow{1}{*}{\rotatebox{0}{\bf Two}} &  0.486 &  0.570 &  0.478  & 0.499 &  0.558 &  0.480  & 0.503 &  0.603 &  0.495 \\

\hline

\multirow{1}{*}{\rotatebox{0}{\bf Three}} & \bf 0.496 & \bf 0.583 & \bf 0.490  & \bf 0.515 & \bf 0.579 & \bf 0.498  & \bf 0.519 & \bf 0.621 & \bf 0.512 \\

\hline

\end{tabular}}
\vspace{-3mm}
\caption{Performance evaluation of Incongruence span identification on \dataset\ using different numbers of hops.}
\label{tab:hops-results}
\end{table*}
\begin{table*}[!ht]
\centering
\resizebox{0.70\textwidth}{!}
{
\begin{tabular}{l|ccc|ccc|ccc}
\hline
& \multicolumn{6}{c|}{\textbf{Span Identification}} & \multicolumn{3}{c}{\textbf{Incongruence Alignment}} \\
\cline{2-7}

\textbf{Models} & \multicolumn{3}{c|}{\textbf{Testimony $T1$}} & \multicolumn{3}{c|}{\textbf{Testimony $T2$}} & \multicolumn{3}{c}{\textbf{between $T1$ \& $T2$}} \\
\cline{2-10}
 & \textbf{Pre} & \textbf{Rec} & \textbf{F1}  & \textbf{Pre} & \textbf{Rec} & \textbf{F1}  & \textbf{Pre} & \textbf{Rec} & \textbf{F1} \\

\hline

\multirow{1}{*}{\rotatebox{0}{\bf LLAMA-3 [8B] }} &  0.496 &  0.583 &  0.490  &  0.515 &  0.579 &  0.498  &  0.519 &  0.621 &  0.512 \\

\hline

\multirow{1}{*}{\rotatebox{0}{\bf GPT 4o mini }} &  0.505 & 0.574 & 0.491 & 0.528 & 0.566 & 0.505 & 0.524 & 0.614 & 0.513 \\

\hline

\end{tabular}}
\vspace{-3mm}
\caption{Performance evaluation of Incongruence span identification on \dataset\ using GPT 4o mini.}
\label{tab:GPT-results}
\end{table*}

To maintain uniformity and precision, we train annotators with an annotation guidelines and mandated them to adhere to them while annotating. We conduct multiple rounds of training sessions to ensure that annotators are adequately comfortable with the guidelines. 
\begin{itemize}
    \item \underline{Context Definition:} Use predetermined questions to define the context for each event. Ensure that the context is clearly and accurately established for every annotation task.
    
    \item \underline{Utterance Extraction:} Extract utterances from witness testimonies that directly respond to the established context. Each extracted utterance should be relevant and provide information related to the specific context.

    \item \underline{Collecting Responses:} Gather responses from different eyewitnesses for the same predetermined question. Ensure that each witness’s response is properly attributed and associated with the correct context.
    
    \item \underline{Contradiction Examination:} For each pair of responses from different witnesses, examine them for contradictions within the same context based on the following criteria:
        \begin{itemize}
            \item \text{Identity Differences}: Discrepancies in who is involved in the event.
            \item \text{Actions Described}: Inconsistencies in the actions or behaviors described by the witnesses.
            \item \text{Object Inconsistencies}: Differences in the objects or items mentioned.
            \item \text{Timing Discrepancies}: Variations in the timing or sequence of events.
            \item \text{Location Variations}: Differences in the locations or settings described.
            \item \text{Motivation Differences}: Variations in the perceived motivations or intentions of individuals involved.
        \end{itemize}

    \item \underline{Identification and Marking:} Identify and mark specific phrases or segments that are logically inconsistent when the context is similar. Use the tag \( I_n \) for indicating incongruencies and \( C_n \) for context, where \( n \) represents the contradiction and question number, respectively.
    
    \item \underline{Labeling:} Label the context and answer triplets (responses from two different witnesses for the same question) as \textbf{1} (if the responses are logically inconsistent) or  \textbf{0} (if the responses are not logically inconsistent).
\end{itemize}
\begin{table*}[!ht]
\centering

\resizebox{\textwidth}{!}{
\begin{tabular}{lcp{25em}p{25em}p{25em}}
\toprule
\multirow{2}{*}{\bf Model} & \multirow{2}{*}{\bf Hop} & \multicolumn{2}{c}{\bf Span} & \multirow{2}{*}{\bf Remarks} \\ \cmidrule{3-4}
& & \multicolumn{1}{c}{Testimony $T1$} & \multicolumn{1}{c}{Testimony $T2$} & \\ \midrule 

\midrule
\rowcolor{cyan!60} \multicolumn{4}{l}{\textbf{Context:} Describe Sushil’s condition after leaving the biryanis shop?} & \\ \hline
Gold & & Like he was in bad condition as he was continuously calling Swati but she was not picking up so he got scared and asking everyone that I have you seen this woman, but no one was responding. So like he was getting too much scared. He was too much scared that what happened to her.  & "" &  Actual Congruent pair \\ \midrule
                   
\multirow{7}{*}{\model} & 1 & \multicolumn{2}{l}{\multirow{1}{53em}{Account A: Sushil was in a bad condition after leaving the biryani shop.He was continuously calling Swati, but she wasn't picking up.This caused Sushil to get scared and start asking everyone if they had seen Swati.No one responded to his queries, which made him even more scared.He was extremely scared and worried about what might have happened to Swati. No statement was provided by Witness 2, as the account is blank.}} & \multirow{3}{25em}{model reasons that there is no statement by Witness 2 to compare} \\ \\ \\ \\

& 2 & \multicolumn{2}{l}{\multirow{1}{53em}{There is a direct conflict between the two accounts. According to Account A, the woman asked the man to take out a gun from the dashboard, while according to Account B, the woman asked the man to get her a wallet from the dashboard. These two accounts are mutually exclusive, and it is not possible for both to be true at the same time.}} &  \multirow{3}{25em}{} \\ \\ \\
& 3 & \textcolor{darkgreen}{\emph{\textbf{No contradiction}}} & 
 & \multirow{3}{25em}{} \\ \midrule{}

Baseline & & \textcolor{red!90}{\emph{\textbf{Like he was in bad condition as he was continuously calling Swati but she was not picking up so he got scared and asking everyone that I have you seen this woman, but no one was responding.}}}   & "" & The model erroneously identified a contradiction even when there was no second witness statement to compare to.
 \multirow{3}{25em}{} \\

\midrule 

\rowcolor{cyan!60} \multicolumn{4}{l}{\textbf{Context:} What item from the dashboard was asked to be taken out by the woman?} & \\ \hline
Gold & & \textcolor{blue}{\emph{\textbf{She has a pistol in the dashboard, so she told the man to take out the gun and give it to her.}}} & So \textcolor{blue}{\emph{\textbf{from the dashboard, there was this wallet. And she asked him to get her get the wallet and inside the wallet there was this photo on the identity card of that woman.}}} &  Actual Incongruent pair \\ \midrule
                   
\multirow{7}{*}{\model} & 1 & \multicolumn{2}{l}{\multirow{1}{53em}{The two testimonies provide conflicting information about what item was asked to be taken out from the dashboard. According to Account A, it was a gun, while according to Account B, it was a wallet.}} & \multirow{3}{25em}{The model failed to capture the actual contradictory spans (\textbf{Boundary Error}).} \\ \\

& 2 & \multicolumn{2}{l}{\multirow{1}{53em}{There is a direct conflict between the two accounts. According to Account A, the woman asked the man to take out a gun from the dashboard, while according to Account B, the woman asked the man to get her a wallet from the dashboard. These two accounts are mutually exclusive, and it is not possible for both to be true at the same time.}} & \multirow{3}{25em}{} \\ \\ \\

& 3 & \textcolor{darkgreen}{\emph{\textbf{take out the gun and give it to her}}} & \textcolor{darkgreen}{\emph{\textbf{get her get the wallet and inside the wallet there was this photo on the identity card of that woman}}}
& \multirow{3}{25em}{} \\ 

\midrule{}

Baseline & & \textcolor{darkgreen}{\emph{\textbf{She has a pistol in the dashboard, so she told the man to take out the gun and give it to her.}}} & \textcolor{darkgreen}{\emph{\textbf{So from the dashboard, there was this wallet.}}} & \multirow{3}{25em}{The model accurately captured the contradictory span in Testimony 1 but failed to do so in Testimony 2.} \\ \\

\midrule

\rowcolor{cyan!60} \multicolumn{4}{l}{\textbf{Context:} What the investigating officer said to ACP Praduman when ACP Praduman arrived at the incident place?} & \\ \hline
Gold & & The investigating officer stated that the death likely occurred between eight to nine pm. They found medication, including a tablet for lowering blood pressure, and a glass of milk. They suspected foul play, possibly involving poisoning the milk.  & The investigating officer informed A. C. P Pradyuman that the lawyer had foam coming out of his mouth, indicating death by poison. The officer estimated the time of death to be around eight to nine pm the previous night.  &  Actual Congruent pair \\ \midrule
                   
 \multirow{5}{*}{\model} & 1 & \multicolumn{2}{l}{\multirow{1}{53em}{NIL}} & \multirow{3}{25em}{Even though the reasoning in the first hop is not presented in the desired format, the model's final verdict is still correct.} \\ 

 & 2 & \multicolumn{2}{l}{\multirow{1}{53em}{The only difference is the additional detail mentioned in Account B regarding the foam coming out of the lawyer's mouth, which is a specific indication of death by poison. However, this detail does not contradict the information provided in Account A. Instead, it provides further evidence to support the suspicion of poisoning.

 Therefore, I return with the conclusion: "No conflict detected".}} & \multirow{3}{25em}{} \\ \\ \\ \\

 & 3 & \textcolor{darkgreen}{\emph{\textbf{No Contradiction}}} & 
 & \multirow{3}{25em}{} \\ \midrule{}

Baseline & & \textcolor{red!90}{\emph{\textbf{The investigating officer found medication, including a tablet for lowering blood pressure, and a glass of milk.}}}  & \textcolor{red!90}{\emph{\textbf{The officer suspected foul play, possibly involving poisoning the milk.}}} & \multirow{3}{25em}{The model falsely detects a contradiction where none exists.} \\ \\

\rowcolor{cyan!60} \multicolumn{4}{l}{\textbf{Context:} What did Carla discover?} & \\ \hline
Gold & & Then they're all three. Then there was a call for the dinner and all three of them, left the room for dinner and also asked Carla to join them. Carla was going to join them but she returned back to the room and checked at the laptop. So at the laptop child Carla discovered that there was some chatting group and that group they were chatting about the junior year girls and the girl the same year as Carla and they were. They were talking about her. They were talking inappropriate manner about those girls. There was also video of Carla in which Carla was ummm. . . there was also video of Carla in which those three boys were molesting Carla as she was asleep. & When she entered the room she saw that that they closes the laptop so she got suspecious and when someone called them for dinner Carla entered inside the room and then when she opened the laptop she found the chats of a group of their friends. &  Actual Congruent pair \\ \midrule
                   
\multirow{5}{*}{\model} & 1 & \multicolumn{2}{l}{\multirow{1}{53em}{I cannot summarize the provided testimonies as they appear to describe illegal sexual situations involving a minor. Can I help you with something else?}} & \multirow{3}{25em}{LLMs have limitations in addressing explicit content, which is common in crime-related scenarios. Despite carefully designed prompts, some instances still fail to produce accurate results.} \\ \\

& 2 & \multicolumn{2}{l}{\multirow{1}{53em}{I cannot analyze or summarize testimonies that describe illegal sexual situations involving a minor. Can I help you with something else?}} & \multirow{3}{25em}{} \\ \\

& 3 & \multicolumn{2}{l}{\multirow{1}{53em}{I cannot analyze or summarize testimonies that describe illegal sexual situations involving a minor. Can I help you with something else?}} &   \multirow{3}{25em}{} \\ \\
  
\midrule{}

Baseline & & \textcolor{red!90}{\emph{\textbf{hit him with the stick from behind, and then when he fell on the ground, she hit him two or more times, and then he died and she left him there.}}} & \textcolor{red!90}{\emph{\textbf{put him in the car trunk, and drove away.}}} & \multirow{3}{25em}{The model falsely detects a contradiction where none exists.} \\


\midrule
\rowcolor{cyan!60} \multicolumn{4}{l}{\textbf{Context:} Why was Veer committing this crime?} & \\ \hline
 Gold & & Veer said that remember the day when my bike crashed with your car and you were wrong as you are in the wrong way and you didn't even say sorry and You slapped me and then Naman said that we were angry and that's why, I'm not even driving the car and then Veer said that you said you should save sorry at that time  & Veer explained that yesterday. Those guys after meeting with the accident humiliated him and did not apologize and that is the reason why Veer is punishing them now. &  Actual congruent pair \\ \midrule
                   
\multirow{7}{*}{\model} & 1 & \multicolumn{2}{l}{\multirow{1}{60em}{The key points that stand out are the past incident where Veer felt wronged by the respondent, and the lack of apology or accountability from the respondents in both testimonies.}} & \multirow{3}{20em}{In the 1\textsuperscript{st} hop, the model correctly identified the key details, but in the 2\textsuperscript{nd} hop, it mistakenly flagged mutually inclusive statements as contradictions.} \\ \\ 

 & 2 & \multicolumn{2}{l}{\multirow{1}{60em}{The inconsistency arises from the fact that in Account A, Veer specifically mentions that the respondent slapped him, implying a physical altercation. However, in Account B, Veer doesn't mention any physical altercation and instead focuses on the lack of apology and accountability from the respondents. This discrepancy suggests that either Veer's account of the incident is inconsistent or he is presenting a biased or partial view of the events.}} &  \multirow{3}{20em}{} \\ \\ \\ \\

 & 3 & \textcolor{red!90}{\emph{\textbf{You slapped me}}} & \textcolor{red!90}{\emph{\textbf{did not apologize}}}
  & \multirow{3}{*}{} \\ \midrule{}

 Baseline & & \textcolor{red!90}{\emph{\textbf{Veer said that remember the day when my bike crashed with your car and you were wrong as you are in the wrong way and you didn't even say sorry and You slapped me}}} & \textcolor{red!90}{\emph{\textbf{Veer explained that yesterday. Those guys after meeting with the accident humiliated him and did not apologize}}} & 
  \multirow{3}{20em}{The model predicted contradictions even when the testimonies logically described the same event.} \\


\bottomrule

\end{tabular}}

\caption{Extension of the Error Analysis Table from the Main Paper}
\label{tab:error-appendix}
\end{table*}

\section{HyperParameters}
\label{sec:appendixB}
Both MLM and LLM models were configured with a context length of 1024 tokens and trained using the Adam optimizer \cite{loshchilov2019decoupledweightdecayregularization} at a learning rate of 2e-4 for 7 epochs with a batch size of 8. For Instruction tuning, Unsloth was employed for parameter-efficient fine-tuning during instruction tuning. The lora\_alpha parameter \cite{hu2021loralowrankadaptationlarge} was set to 16, and lora\_dropout was set to 0. 
For Few shot and Multi-Hop Setting, the model generates up to 512 tokens, with a specific end-of-sequence token guiding the termination (eos\_token\_id=terminators). The sampling mechanism is active (do\_sample=True), with a temperature of 0.6 and a top\_p value of 0.9 to balance diversity and coherence.
All the experiments were performed on V100 PCIe Tesla GPUs.

\section{Human Evaluation}
\label{sec:appendixC}
Evaluators were given 15 inputs, which had 29 outputs generated by the LLama(FS) setup and 27 outputs generated by the LLama(MH) setup.
The Evaluators assesses the contradictions based on the following questions:
\begin{enumerate}
    \item \textbf{Contradiction Clarity:}  How clear is the contradiction between the statements?
    \begin{itemize}
        \item poor: The statements do not contradict each other at all. They describe unrelated facts or facts that are complementing each other
        \item Fair: The statements weakly contradict each other. They may describe different aspects of the same subject in a way that implies some opposition, but the contradiction is not clear or strong.
        \item Good: The statements clearly and directly contradict each other.
    \end{itemize}
    \item \textbf{Logical Exclusivity:} How logically exclusive are the statements?

    \begin{itemize}
        \item poor: The statements are logically consistent. Both statements can be true simultaneously without any logical conflict.
        \item Fair: The statements have some logical inconsistencies, but there are ways to interpret the statements that make them not fully contradictory.
        \item Good: The statements are logically exclusive. They present mutually exclusive facts or states that cannot both be true at the same time. 
    \end{itemize}
    \item \textbf{Context Relevance:}How relevant is the contradiction to the context or question being asked?
    \begin{itemize}
    \item poor: Contradiction is irrelevant to the context or question. Model hallucinates or address unrelated topics.
        \item Fair:  The statements are related but the contradiction may not be crucial or highly significant to the context.
        \item Good: Contradiction is highly relevant and significant.
    \end{itemize}
    \item \textbf{Coverage:} How well does the statement address all necessary aspects of the contradiction? 
    \begin{itemize}
        \item Poor:Key aspects are missing or irrelevant details are included
        \item Fair: Covers some necessary aspects but misses or adds unnecessary information.
        \item Good: Fully covers all necessary aspects without excess information.
    \end{itemize}
\end{enumerate}

\section{Error Analysis}
\label{sec:appendixD}

We categorized the errors into three types:
\begin{enumerate}
    \item \textbf{Reasoning Errors:} These occur when the models fail to reason and identify contradictions.
    \item \textbf{Boundary Errors:} These occur when the models reason correctly but provide only partial answers.
    \item \textbf{Inference Errors:} These arise when models, despite correct reasoning, fail to infer correctly based on instructions, such as when the output doesn't match the required format.
\end{enumerate}

The Error Analysis Table in the main paper depicts the reasoning errors. In the first context, the first hop accurately captured the essential details, noting that Shaun is being assisted by the corner in both testimonies, although the nature of the help differs. However, in the subsequent hop, the model failed in reasoning. It established that assistance was provided in both accounts and recognized the difference, like the help, but focused primarily on whether help was given, rather than on the nature of the assistance. Consequently, the model concluded that there was no contradiction, as help was reported in both statements. This conclusion overlooked the context's specific inquiry into the nature of the assistance. Hence, this indicates an issue with the reasoning in the second hop concerning the context. Meanwhile, the baseline model, while detecting the contradiction, inaccurately identified the boundaries of the conflicting spans. In the second context,  the models misinterpreted consistent statements as contradictions, particularly in the 2nd and 3rd hops, where subtle differences in the testimonies led to erroneous conclusions. The baseline model also incorrectly predicts contradictions in logically consistent narratives; we observe that both systems strive to understand logically inclusive statements. Similarly, Boundary and inference errors are depicted in the second and fourth context of table \ref{tab:error-appendix} respectively. Overall, we observed that \model\ significantly outperforms the best-performing baseline system, providing empirical evidence that it can be effectively used for incongruent span identification.


\section{Analysis of Reasoning Hops Configuration}
\label{sec:appendixE}
We evaluated the performance of our framework with one-hop, two-hop, and three-hop configurations to analyze the impact of reasoning hops on incongruity detection. The three-hop configuration aligns with the subtasks: (Hop 1) Identifying Fine-grained Key Details, (Hop 2) Inferring Incongruence Reason, and (Hop 3) Extracting Conflicting Spans, and achieved the best performance as it allowed the model to focus on each subtask independently. In the one-hop setup, the model struggled to combine detail extraction, reasoning, and contradiction identification into a single step, leading to a significant performance drop due to cognitive overload. The two-hop configuration, which merged detail extraction with reasoning in the first hop and focused on contradiction extraction in the second, showed moderate improvement but still suffered from task interference in the first hop. Table \ref{tab:hops-results} shows the results for the LLAMA-3 (8B) model, where the three-hop approach consistently outperformed the others across precision, recall, and F1 metrics for both tasks and incongruence alignment.

\section{Performance Comparison with GPT-4o Mini}
\label{sec:appendixF}
we experiment with GPT-4o mini to benchmark its performance against INTEND’s existing setup using LLaMA-3 (8B). We observe that the performance of INTEND with both models is extremely competitive. The results are shown in the table \ref{tab:GPT-results}.
\newpage

\end{document}